\documentclass[letterpaper]{article} 
\usepackage{aaai23}  
\usepackage{times}  
\usepackage{helvet}  
\usepackage{courier}  
\usepackage[hyphens]{url}  
\usepackage{graphicx} 
\usepackage{natbib}  
\usepackage{caption} 
\usepackage[lined,ruled,vlined,commentsnumbered]{algorithm2e}
\usepackage{graphicx}
\usepackage{amsmath}
\usepackage{amssymb}
\usepackage{booktabs}
\usepackage{times}
\usepackage{helvet}
\usepackage{courier}
\usepackage{epsfig}
\usepackage{mathrsfs}
\usepackage{multirow}
\usepackage{color}
\usepackage{xcolor}
\usepackage{colortbl}
\usepackage{tabularx}
\usepackage{gensymb}
\usepackage{arydshln}
\usepackage{pifont}

\newcommand{\ljx}[1]{\textcolor{black}{#1}}
\newcommand{\seasons}[1]{\textcolor{black}{#1}}

\title{SpatialFormer: Semantic and Target Aware Attentions for Few-Shot Learning}
\author {
    Jinxiang Lai \textsuperscript{\rm 1},
    Siqian Yang \textsuperscript{\rm 1},
    Wenlong Wu \textsuperscript{\rm 1},
    Tao Wu \textsuperscript{\rm 1},
    Guannan Jiang \textsuperscript{\rm 2},
    Xi Wang \textsuperscript{\rm 2},
    Jun Liu \textsuperscript{\rm 1},
    Bin-Bin Gao \textsuperscript{\rm 1},
    Wei Zhang \textsuperscript{\rm 2},
    Yuan Xie\thanks{Corresponding Author} \textsuperscript{\rm 3},
    Chengjie Wang\textsuperscript{*} \textsuperscript{\rm 1, 4}
}
\affiliations {
    \textsuperscript{\rm 1} Tencent Youtu Lab, China \quad

    \textsuperscript{\rm 2} CATL, China \quad

    \textsuperscript{\rm 3} East China Normal University, China \quad

    \textsuperscript{\rm 4} Shanghai Jiao Tong University, China\\

    \{jinxianglai, seasonsyang, ezrealwu, tobinwu\}@tencent.com, \{jianggn, wangx30, zhangwei\}@catl.com \\
    \{junsenselee, csgaobb\}@gmail.com, yxie@cs.ecnu.edu.cn, jasoncjwang@tencent.com
}

\usepackage{bibentry}

\begin{document}

\maketitle

\begin{abstract}
Recent Few-Shot Learning (FSL) methods put emphasis on generating a discriminative embedding features to precisely measure the similarity between support and query sets.
Current CNN-based cross-attention approaches generate discriminative representations via enhancing the mutually semantic similar regions of support and query pairs.
However, it suffers from two problems: CNN structure produces inaccurate attention map based on local features, and mutually similar backgrounds cause distraction.
To alleviate these problems, we design a novel SpatialFormer structure to generate more accurate attention regions based on global features.
Different from the traditional Transformer modeling intrinsic instance-level similarity which causes accuracy degradation in FSL, our SpatialFormer explores the semantic-level similarity between pair inputs to boost the performance.
Then we derive two specific attention modules, named SpatialFormer Semantic Attention (SFSA) and SpatialFormer Target Attention (SFTA), to enhance the target object regions while reduce the background distraction.
Particularly, SFSA highlights the regions with same semantic information between pair features, and SFTA finds potential foreground object regions of novel feature that are similar to base categories.
Extensive experiments show that our methods are effective and achieve new state-of-the-art results on few-shot classification benchmarks.
\end{abstract}

\section{Introduction}
\label{sec:intro}
Few-Shot Learning (FSL) problem becomes one of the research hotspots due to its special ability that could classify novel (unseen) classes samples relying on few labeled images.
In this field, recent works \cite{hou2019cross,tian2020rethinking,rizve2021exploring,zhengyu2021pareto,zhiqiang2021partial,liu2021learning,xu2021learning} focus on designing a fine-grained embedding network and improving its robustness to recognize the novel instances.

\begin{figure}[!t]
\centering
\includegraphics[width=0.99\linewidth]{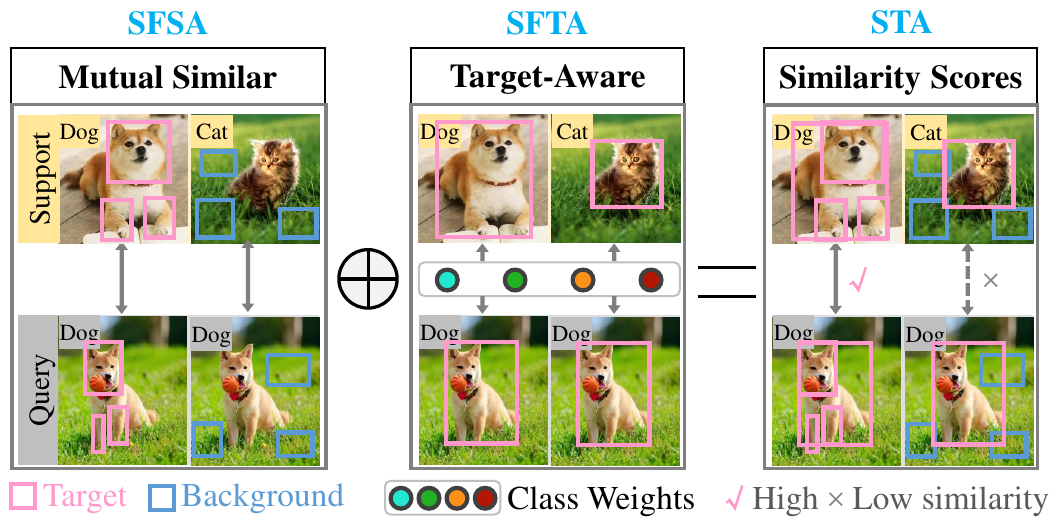}
\caption{As shown in 1st column, the limitation of cross-attention based approaches is that they only recognize the mutual similar object, but can not solve the distraction problem causing by similar backgrounds.
In this paper, we propose two effective modules called SFSA and SFTA, which are utilized to find the mutual similar object and obtain target-aware information, respectively. With their combination, the derived STA module is able to enhance target object while alleviate background distraction.}
\label{fig:motivation}
\end{figure}

The criterion of a fine-grained embedding is the ability of distinguish background and the target object.
\ljx{As presented in 1st column of Fig.~\ref{fig:motivation}, to generate discriminative features, current Convolutional Neural Network (CNN) based cross-attention methods \cite{hou2019cross,xu2021learning} highlight the mutually similar regions of support and query pairs, which assume these similar regions are the target objects.}
\ljx{However, these methods suffer from two problems:}
\begin{itemize}
\item
\ljx{\emph{Problem \ding{172}: The CNN-based cross-attention produces inaccurate attention map due to modeling correlation relation based on local features.}}
\item
\ljx{\emph{Problem \ding{173}: The mutually similar backgrounds cause distraction.}
For example, in Fig. \ref{fig:motivation}, the query contains dog and grass, while the support image belonging to cat category also has grass. So the meta classifier may regard that this query (dog + grass) is more similar to the support image of cat with grass.
}
\end{itemize}

\ljx{To deal with \emph{Problem \ding{172}}, we manage to employ Transformer \cite{vaswani2017attention,dosovitskiy2020image,wang2021pyramid}, which is able to obtain global information, and has the potential ability to generate a more accurate attention region than CNN-based cross-attention module.
However, it raises \emph{Problem \ding{174}: The traditional Transformer recommends the inputs are instance-level similar, it against the requirement of FSL which inputs semantic-level similar query and support pairs.}}
In detailed, SuperGlue \cite{sarlin2020superglue} used Transformer to implement cross-attention between feature pairs for image matching.
But our results in Tab.~\ref{table:ablation_1} indicates that SuperGlue causes obvious performance degradation in FSL.
According to the current applications, Transformer recommends that the input features are instance-level similar.
Specifically, the self-attention based Transformer inputs single feature, and utilizes convolution layers to generate perturbed different features.
The cross-attention based Transformer, e.g. SuperGlue, inputs pair features that are generated from same instance objects.
Obviously, in few-shot classification, the relationship of input support and query pairs violates the recommendation of instance-level similar, i.e. the pairs from same category are semantic-level similar.

\begin{figure}[!t]
\centering
\includegraphics[width=0.99\linewidth]{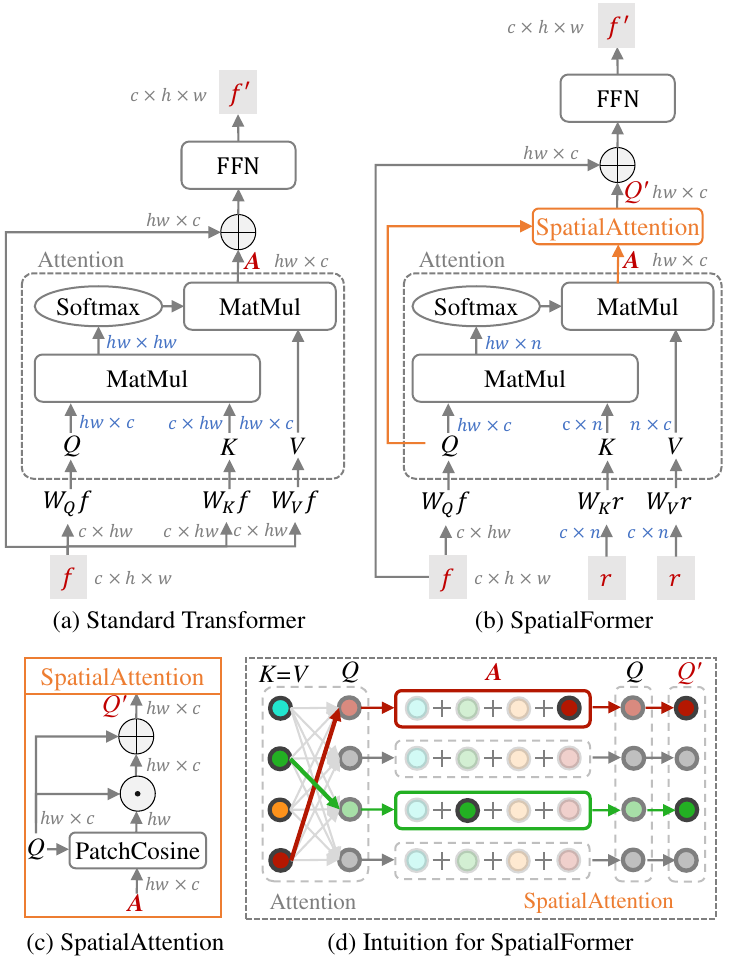}
\caption{(a) shows a standard Transformer structure.
(b) is the proposed SpatialFormer, where ${(f,r)}$ are input feature and \emph{reference object}, and SpatialAttention operation is presented in (c). (d) gives the intuition for SpatialFormer, which uses ${A}$ (i.e. ${Q}$-aligned ${V}$) to strength the spatial response of ${Q}$.
\ljx{And the dimension of each node (i.e. the circle in (d)) is $\mathbb{R}^{c}$, which denotes each spatial position of features. The gray arrow represents low similarity relationship, and the colorful arrow is high similarity.}
}
\label{fig:SF}
\end{figure}

\ljx{To solve \emph{Problem \ding{174}}, as illustrated in Fig.~\ref{fig:SF}, we design a novel variant Transformer structure called SpatialFormer to perform spatial-wise cross-attention, which allows the inputs to be semantic-level similar.}
Our SpatialFormer enhances the spatial response of the input features via the similar regions with \emph{reference object}. The \emph{reference object} can be an instance feature or other forms that has semantic relations with the input features.
In practice, given input pair features, SpatialFormer can highlight the mutually similar regions of them with the same semantic information, which is called as SpatialFormer Semantic Attention (SFSA).

To alleviate \emph{Problem \ding{173}}, we propose two effective modules called SpatialFormer Target Attention (SFTA) and Novel Task Attention (NTA) to enhance target object while reduce the background distraction.
Specifically, as shown in Fig. \ref{fig:motivation}, given input feature, SFTA assigns the class weights (i.e. the weights of linear classifier learned on base dataset) as the \emph{reference object}, to highlight the potential foreground object regions. Particularly, the semantic information of base dataset is embedded into the class weights, which represents base target-aware information. Benefiting from SpatialFormer, SFTA utilizes the class weights to enhance the input novel feature. Besides, NTA involves task-aware information to increase the inter-class feature distance for distinguishing different categories.

In general, our contributions are listed as follows:
\begin{itemize}
\item
A novel SpatialFormer structure is designed for few-shot learning, which explores semantic-level similar between inputs. It gets rid of intrinsic instance-level similarity in Transformer, which causes accuracy degradation in FSL.
\item
A Semantic and Target Attentions (STA) module is proposed to enhance the target object helping the subsequent metric classifier to precisely measure similarity. The STA consists of SpatialFormer Semantic Attention (SFSA) and SpatialFormer Target Attention (SFTA), which are utilized to find the mutual similar object and obtain target-aware information, respectively.  Besides, for the first time, our SFTA proves that the class weights learned from the base dataset are useful in few-shot learning.
\item
A Novel Task Attention (NTA) module is introduced to increase the inter-class feature distance via involving task-aware information, which is helpful to generate a more discriminative embedding features.
\item
Based on STA and NTA modules, a novel STANet framewrok is derived for few-shot classification.
Extensive experiments show that our STANet achieves new state-of-the-art results on few-shot classification benchmarks such as miniImageNet and tieredImageNet.
\end{itemize}

\section{Related Work}
\label{sec:related_work}
\noindent\textbf{Few-Shot Learning} \
FSL algorithms aim to classify novel classes samples using only few labeled images, and a base dataset containing sufficient labeled images is provided for model pre-training.
Four main kinds of FSL algorithms are briefly summarized in below.
\emph{Optimization-based} approaches achieve fast model adaptation with few novel images \cite{ravi2016optimization,rusu2019meta}.
\emph{Parameter-generating} approaches \cite{bertinetto2016learning,gidaris2019generating} design a parameter generating network.
\emph{Embedding-based} approaches \cite{tian2020rethinking,rizve2021exploring} pay more attention to improve the generalization ability of embedding.
\emph{Metric-based} approaches perform classification via calculating similarity between a given input image and the labeled classes samples \cite{vinyals2016matching,snell2017prototypical}.
\ljx{Besides, some metric-based approaches (e.g. CAN \cite{hou2019cross} amd DANet \cite{xu2021learning}) integrate attention modules to generate discriminative embedding.}
However, these CNN-based approaches produce the correlation maps based on local features, without global information embedding.


\noindent\textbf{Auxiliary Task for FSL} \
The auxiliary tasks performing supervised \cite{hou2019cross} or self-supervised \cite{liu2021learning,lai2022tsf,lai2022rethinking} learning are widely used to improve the generalization ability of few-shot models.

\noindent\textbf{Transformer for FSL} \
\ljx{The FEAT \cite{ye2020feat} and CTX \cite{doersch2020crosstransformers} apply standard Transformer \cite{vaswani2017attention} structure in feature embedding for FSL.
These two methods do not highlight the mutually similar regions between support and query features at semantic level, i.e. they are not able to handle \emph{Problem \ding{172}}.}

\section{SpatialFormer}
\subsection{\seasons{Traditional} Transformer}
\noindent\textbf{Transformer Self-Attention} \
As shown in Fig.~\ref{fig:SF}(a), a standard Transformer \cite{vaswani2017attention} contains an Attention Module and follows a Feed-Forward Network (FFN).
\seasons{Let ${f \in \mathbb{R}^{c\times h\times w}}$ and ${f{'} \in \mathbb{R}^{c\times h\times w}}$ denote the input and output features of a traditional Transformer, respectively.}
Formally,
\begin{equation}
\begin{aligned}
&f{'} = Transformer(f)= FFN\left(f + A\right),\\
&where, \, A=Attention\left(Q,K,V\right)=\sigma(QK^T)V, \\
&where, \, Q=W_{Q}f,\quad K=W_{K}f,\quad V=W_{V}f,
\label{equ:transformer}
\end{aligned}
\end{equation}
where, $\{W_{Q},W_{K},W_{V}\}$ are convolution layers, and ${\sigma}$ is softmax function. ${\{Q,K,V\} \in \mathbb{R}^{hw \times c }}$ represent $\{\emph{query}, \emph{key}, \emph{value}\}$ as well-known in Transformer.
For few-shot classification task, Feat \cite{ye2020few} performs embedding adaptation via using the Transformer as a set-to-set function, which makes the modifications as:
$Q$=$W_{Q}f_{set}$, $K$=$W_{K}f_{set}$, $V$=$W_{V}f_{set}$,
where ${f_{set}}$ is a feature set of all support samples.
Briefly, the vanilla Transformer and Feat learn the correlation of instance features via using self-attention mechanism.

\noindent\textbf{Transformer Cross-Attention} \
Instead of performing self-attention with Transformer, SuperGlue \cite{sarlin2020superglue} performs Transformer cross-attention on a pair of input features.
Given a pair of input features (${f_q, f_{s}}$), it first re-weights ${f_{s}}$ with the affinity matrix between (${f_q,f_{s}}$), and then the weighted ${f_{s}}$ is used to enhance the input feature $f_q$.
Formally, the ${\{Q,K,V\}}$ of \emph{Transformer cross-attention} are expressed as:
$Q$=$W_{Q}f_{q}$, $K$=$W_{K}f_{s}$, $V$=$W_{V}f_{s}$.

\noindent\textbf{Transformer Alignment} \
For few-shot classification task, CTX \cite{doersch2020crosstransformers} uses the Transformer structure to produce query-aligned prototype ${f_p}$. \seasons{The formulation is listed as follows}:
\begin{equation}
\begin{aligned}
{f_p}&=\sum_{i=1}^{M}\sigma({Q}{K}_i^T)V_i, \\
 where, \quad Q&=W_{Q}f_q,\,\, K_i=W_{K}f_{s}^i,\,\, V_i=W_{V}f_{s}^i,
\label{equ:ctx}
\end{aligned}
\end{equation}
where, $M$ is the number of support \seasons{instances} per category, $f_q$ \seasons{donates} the query feature, and $f_s^i$ is the support feature of $i^{th}$ image. CTX aggregates $M$ support features into alignment with query $f_q$.
\ljx{Thus CTX achieves feature alignment via implementing cross-attention, while does not enhance the target regions of input feature $f_q$.}

\begin{figure*}[!t]
\centering
\includegraphics[width=0.85\linewidth]{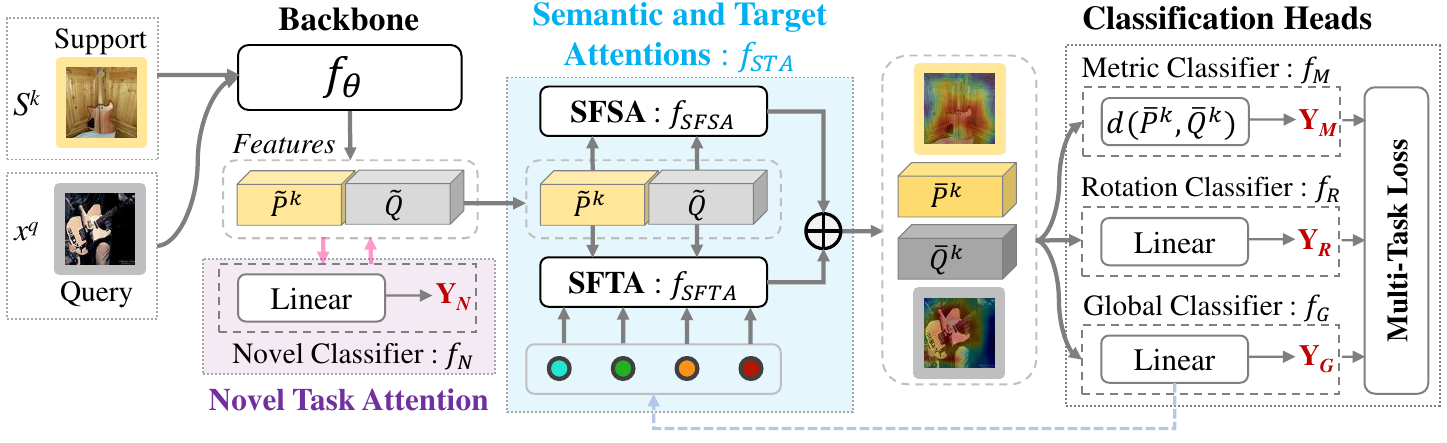}
\caption{
The STANet framework.
(a) The designed Novel Task Attention (NTA) module obtains novel task information via utilizing the weights of Novel Classifier, which is fine-tuned on support data of novel set ${X^{novel}}$.
(b) The proposed Semantic and Target Attentions (STA) module enhances target object regions. Specifically, SpatialFormer Semantic Attention (SFSA) module highlights mutually similar regions between support and query features. \seasons{Meanwhile}, SpatialFormer Target Attention (SFTA) module highlights potential foreground object regions via utilizing base target information embedded in Global Classifier.}
\label{fig:STANet}
\end{figure*}

\subsection{SpatialFormer Methodology}
As illustrated in Fig.~\ref{fig:SF}(b), our SpatialFormer applies \emph{Transformer Spatial-Attention}, which enhances the spatial response of the input features ${f}$ via the similar regions with \emph{reference object} ${r}$.
Given \seasons{a} pair of feature ${f \in \mathbb{R}^{c\times h\times w}}$ and \emph{reference object} ${r \in \mathbb{R}^{c\times n}}$ \seasons{as input}, SpatialFormer outputs the enhanced feature ${f{'} \in \mathbb{R}^{c\times h\times w}}$.
Formally, SpatialFormer is expressed as:
\begin{equation}
\begin{aligned}
&f{'} = SpatialFormer(f,r)= FFN\left(f + Q{'}\right),\\
&where, \, Q{'}=SpatialAttention\left(Q,A\right), \\
&where, \, A=Attention\left(Q,K,V\right)=\sigma(QK^T)V, \\
&where, \, Q=W_{Q}f,\quad K=W_{K}r,\quad V=W_{V}r.
\label{equ:full_SF}
\end{aligned}
\end{equation}
$SpatialAttention$ is defined as:
\begin{equation}
\begin{aligned}
Q{'} &= SpatialAttention(Q,A) \\
&=Q + Q \odot PatchCosine(Q,A),
\label{equ:SpatialAttention}
\end{aligned}
\end{equation}
where ${\odot}$ is Broadcasting Element-wise Product to implement spatial-wise attention.
Particularly, $m^{th}$ spatial position of ${(Q,A) \in \mathbb{R}^{hw\times c}}$ are denoted as the patch features $(Q_m,A_m) \in \mathbb{R}^{c}$, then $PatchCosine$ operation calculates the cosine similarity between $(Q_m,A_m)$ for each spatial position of $(Q,A)$, which obtains the result ${PatchCosine(Q,A) \in \mathbb{R}^{hw}}$.

Next, we discuss the meanings of some key variables in SpatialFormer:
(I) $(Q,K,V)$, are corresponding to inputs $(f,r,r)$ processed by convolution layers for feature adjustment, which are denoted as $(Q,K,V) \approx (f,r,r)$.
(II) $A$, is $Q$-aligned $V$. According to the analysis of CTX approach and Fig.~\ref{fig:SF}(d), at $m^{th}$ spatial position, $A_m$ is aggregated from $V$ with the affinity vector between ${(Q_m,K)}$. Intuitively, $A_m$ collects all the patch features from $V$ that are semantically similar to the reference ${Q_m}$.
(III) $Q{'}$, is $A$-enhanced $Q$. $SpatialAttention$ operation uses the cosine similarity score between $({Q_m},{A_m})$ to re-weight $Q_m$ to generate $Q{'}_m$.
Therefore, ${Q{'}_m}$ will be enhanced if ${Q_m}$ is highly similar to ${A_m}$ (i.e. $Q$-aligned $V$), i.e. ${Q_m} \approx f_m$ can be enhanced to produce $Q{'}_m$ by the similar regions with $V \approx r$.
(IV) $f{'}$, is $Q{'}$-enhanced $f$.
Thus SpatialFormer can enhance the spatial response of the input features $f$ by the similar regions with \emph{reference object} ${r}$. The \emph{reference object} ${r}$ can be an instance feature or other forms that has semantic relations with the input features.

According to the definition of SpatialFormer in Eq.~\ref{equ:full_SF}, there is $f{'}=FFN(f+Q{'})$.
Let $f$ and $r$ represent two different instance-level objects.
$Q{'}$ is produced via re-weighting ${Q} \approx f$, thus $Q{'}$ and $f$ are belong to the same instance-level, which is denoted as $Q{'} \sim f$.
Similarly, according to the analysis of (II), there is $A \sim r$.
Comparing to our SpatialFormer, traditional Transformer is defined in Eq.~\ref{equ:transformer} as $f{'}=FFN(f+A)$.
Obviously, the main difference between SpatialFormer and Transformer is from $Q{'} \sim f$ and $A \sim r$.
Transformer updates feature with $FFN(f+A)|A \sim r$, hence the fusion of different instance-level objects may cause serious disturbance on $f$. Therefore, in current applications such as supervised classification and image matching, the inputs of Transformer are instance-level similar.
Differently, our SpatialFormer enhances feature with $FFN(f+Q{'})|Q{'} \sim f$, thus the combination of same instance-level objects does not affect the intrinsic semantic essence of $f$. Therefore, SpatialFormer only requires that the inputs are semantic-level similar, which meets the requirement of few-shot classification task (i.e. the pairs from same category are semantic-level similar).

\section{Semantic and Target Attentions Network}
\subsection{Problem Definition}
\label{sec:ProblemDef}
A ${N}$-way ${M}$-shot few-shot classification task, aims to train a classification model to recognize $N$ novel categories based on $M$ labeled images per class.
Usually, two category-disjoint datasets are provided: a base dataset ${X^{base}}$ with ${C^{base}}$ categories for training, and a novel dataset ${X^{novel}}$ with ${C^{novel}}$ categories for testing, where ${C^{base} \cap C^{novel} = \emptyset}$.
Following \cite{vinyals2016matching,hou2019cross,xu2021learning}, we adopt episodic training paradigm to mimic FSL setting.
The episodic training paradigm samples an episode to mimic an individual FSL task.
An episode consists of a support set $\mathcal{S}=\{\left(x^s_i, y^s_i\right)\}_{i=1}^{m_s}$ ($m_s=N\times M$) and a query set $\mathcal{Q}=\{\left(x^q_i, y^q_i\right)\}_{i=1}^{m_q}$, where ${m_s}$ and ${m_q}$ are the amount of samples of support set and query set, respectively.
Moreover, we denote the $k^{th}$ class support subset as $\mathcal{S}^k$.

\subsection{Method Framework}
As illustrated in Fig.~\ref{fig:STANet}, the proposed Semantic and Target Attentions Network (STANet) contains six parts: embedding backbone ${f_\theta}$, Semantic and Target Attentions (STA) ${f_{STA}}$ module, Metric ${f_M}$ and Novel ${f_N}$ few-shot classifiers, and auxiliary Rotation ${f_R}$ and Global ${f_G}$ classifiers. The proposed STA generates more discriminative representations by highlighting the target object regions of semantic features which benefits the subsequent classifiers. The Novel Task Attention (NTA) operation implemented in Novel Classifier ${f_N}$ obtains novel task information to increase the inter-class feature distance.
our STANet can be divided into two procedures: \textbf{Step 1}, train the base model ${f_{base}={[f_\theta,f_{STA},f_M,f_G,f_R]}}$ on the base set ${X^{base}}$; \textbf{Step 2}, fine-tune the Novel Classifier ${f_N}$ and make predictions with ${f_{MN}={[f_\theta,f_{STA},f_M,f_N]}}$ on the novel set ${X^{novel}}$.

In \textbf{Step 1}, the input image $x^q$ in query set $\mathcal{Q}=\{\left(x^q_i, y^q_i\right)\}_{i=1}^{m_q}$ is rotated with [0\degree, 90\degree, 180\degree, 270\degree] and outputs a rotated $\mathcal{\tilde{Q}}=\{\left(\tilde{x}^q_i, \tilde{y}^q_i\right)\}_{i=1}^{{m_q}\times4}$.
The rotated query $\tilde{x}^q$ and support subset $\mathcal{S}^k$ are processed by the embedding backbone ${f_\theta}$, which generates the query feature $\tilde{Q}={f_\theta}(\tilde{x}^q)\in \mathbb{R}^{c\times h\times w}$ and support prototype $\tilde{P}^k=\frac{1}{|\mathcal{S}^k|} \sum_{x^s_i\in \mathcal{S}^k} {f_\theta}(x^s_i)$, respectively.
Then each pair-features ($\tilde{P}^k$, $\tilde{Q}$) are processed by STA to enhance the target object regions and generates more discriminative features ($\bar{P}^k$, $\bar{Q}^k$) for the subsequent classification.
Finally, ${f_{base}}$ is optimized by Multi-Task Loss defined in Eq.~\ref{equ:Loss}. In \textbf{Step 2}, based on NTA operation, firstly the Novel Classifier ${f_N}$ is fine-tuned on support data of ${X^{novel}}$ via optimizing $\mathcal{L}_N=CE(Y_N,N)$, where $CE$ is cross-entropy function and ${Y_N}$ is the prediction result of Novel Classifier. Then the output embedding features of ${f_\theta}$ is updated by NTA for all input support and query images. Finally the fusion model ${f_{MN}}$ makes predictions on the updated embedding features.
Specifically, in inductive inference phase, the overall prediction of STANet is ${Y}={Y_M}+{Y_N}$, where ${Y_M}$ is the result of Metric classifier.

In the following context, we first describe two novel components STA and NTA, then the adopted Multi-Task Loss is introduced.
The detailed algorithm of STANet is presented in APPENDIX.

\subsection{Semantic and Target Attentions}
The STA module $f_{STA}$ is able to enhance the target object, and consists of SpatialFormer Semantic Attention (SFSA) $f_{SFSA}$ and SpatialFormer Target Attention (SFTA) $f_{SFTA}$, which are utilized to find the mutual similar object and obtain target-aware information, respectively.
The corresponding intuitions of them are presented in Fig. \ref{fig:motivation}.
Formally, STA module is expressed as:
\begin{small}
\begin{equation}
\begin{aligned}
(\bar{P}^k,\bar{Q}^k) &= f_{STA}(\tilde{P}^k,\tilde{Q}) \\
&= f_{SFSA}(\tilde{P}^k,\tilde{Q}) + f_{SFTA}(\tilde{P}^k,\tilde{Q}).
\label{equ:STA}
\end{aligned}
\end{equation}
\end{small}

\noindent\textbf{SpatialFormer Semantic Attention} \
$f_{SFSA}$ is expressed as:
\begin{small}
\begin{equation}
\begin{aligned}
&(\tilde{P}^{k'},\tilde{Q}^{k'}) = f_{SFSA}(\tilde{P}^k,\tilde{Q}) \\
&= SpatialFormer(\tilde{P}^k,\tilde{Q}),\,SpatialFormer(\tilde{Q},\tilde{P}^k)
\label{equ:SFSA}
\end{aligned}
\end{equation}
\end{small}Based on SpatialFormer, SFSA module is able to highlight mutually similar regions between support and query features $(\tilde{P}^k,\tilde{Q})$.
For example, $\tilde{Q}^{k'}$=$SpatialFormer(\tilde{Q},\tilde{P}^k)$: as defined in Eq.~\ref{equ:full_SF} and shown in Fig.~\ref{fig:SF}(b), SpatialFormer enhances the spatial response of  $\tilde{Q}$ via the similar regions with \emph{reference object} $\tilde{P}^k$.

Comaring to CNN-based cross-attention, our SFSA produces more complete attention maps by ${SpatialAttention}$ operation.
Additionally, inheriting Transformer's property of globally exploring correlations among feature instances, SpatialFormer based SFSA can generate more accurate attention maps via modeling whole data set rather than a meta-task done by CNN-based attention.

\noindent\textbf{SpatialFormer Target Attention} \
$f_{SFTA}$ is expressed as:
\begin{small}
\begin{equation}
\begin{aligned}
&(\tilde{P}^{k'},\tilde{Q}^{k'}) = f_{SFTA}(\tilde{P}^k,\tilde{Q}) \\
&= SpatialFormer(\tilde{P}^k,W_G),SpatialFormer(\tilde{Q},W_G)
\label{equ:SFTA}
\end{aligned}
\end{equation}
\end{small}where, $W_G\in \mathbb{R}^{C^{base}\times c}$ is the weights of linear Global Classifier which classifies the input into $C^{base}$ classes of base dataset $X^{base}$.
The proposed SFTA module, utilizes the base class weights $W_G$ as the \emph{reference object}, to highlight the potential foreground regions of the input features.
Intuitively, the semantic information of base dataset $X^{base}$ is embedded into the class weights $W_G$, which represents base target-aware information (i.e. corresponding to $C^{base}$ categories). Benefiting from SpatialFormer, SFTA utilizes $W_G$ to enhance the input novel feature. For the first time, our SFTA shows that the semantic information embedded in the base class weights $W_G$ is useful in few-shot learning.

Formally, let's assume the $C^{base}$ contains two subsets $C^{base}_{sim}$ and $C^{base}_{diff}$, which are semantic similar and different to $C^{novel}$, respectively.
After learning on base dataset $X^{base}$, $W_G\in \mathbb{R}^{C^{base}\times c}$ gathers the semantic info of $C^{base}$, i.e. $W_G$ also contains two subsets $W_{G}^{sim}$ and $W_{G}^{diff}$ that are semantic similar and different to $C^{novel}$, respectively.
Therefore, in testing on $X^{novel}$, our SFTA module can enhance the novel feature regions that are similar to $W_{G}^{sim}$.

\subsection{Novel Task Attention}
\label{subsec:NTA}
\begin{figure}[!t]
\centering
\includegraphics[width=0.9\linewidth]{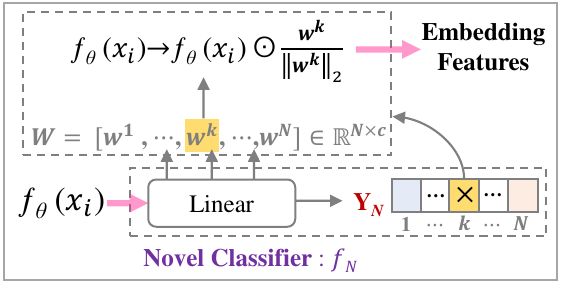}
\caption{The designed Novel Task Attention (NTA) module obtains novel task information via utilizing the weights of Novel Classifier.}
\label{fig:NTA}
\end{figure}

The novel task is sampled from novel set ${X^{novel}}$ for few-shot classification evaluation, which contains a specific task-aware information for each novel task.
As illustrated in Fig.~\ref{fig:NTA}, the Novel Task Attention (NTA) shifts the input feature with the weights of Novel Classifier to generate task-aware feature.

Novel Classifier makes prediction as
$Y_N=W{f_{\theta}(x_i)}$,
where ${W}\in \mathbb{R}^{N\times c}$ are the weights of Linear layer.
After fine-tuning on novel support data, the learned weights ${W}$ of Novel Classifer is utilized to shift the input features. Specifically, the Novel Classifier recognizes the input ${x_i}$ as ${k^{th}}$ category which indicates the strongest response weight is ${w^k \subseteq W}$, thus ${w^k \in \mathbb{R}^{c}}$ is selected as the rectify vector for shifting the embedding feature ${f_{\theta}(x_i)}$. Formally, the updated feature is derived by:
\begin{equation}
{f_{\theta}(x_i)}\leftarrow {f_{\theta}(x_i)} \odot \frac{w^k}{||w^k||_2},
\label{equ:n2b}
\end{equation}
where ${\odot}$ is Broadcasting Element-wise Product to implement channel-wise attention.
The learned ${W}$ represents the response for all classes of novel task, which is considered as task-aware class response, therefore NTA is able to filter out the background noise via the involved task-aware information. Moreover, the features ${f_{\theta}(y=k|x_i)}$ of same category inputs have the same corresponding response weight ${w_k}$, while the response weights of different categories are not the same, thus NTA can increase the inter-class feature distance based on different rectify vectors.

\renewcommand{\tabcolsep}{5.0pt}
\begin{table*}[t]
\centering
\begin{tabular}{ l | c | c c | c c}
\hline
\multicolumn{1}{c|}{\multirow{2}*{model}}  & \multirow{2}*{Backbone} & \multicolumn{2}{c|}{miniImageNet}  &\multicolumn{2}{c}{tieredImageNet} \\
\cline{3-6}
\multicolumn{1}{c|}{ } & & 1-shot &5-shot &1-shot &5-shot \\
\hline
ProtoNet \cite{snell2017prototypical} &Conv4 &49.42 $\pm$ 0.78 & 68.20 $\pm$ 0.66 &53.31 $\pm$ 0.89 &72.69 $\pm$ 0.74\\
RelationNet \cite{sung2018learning} &Conv4 &50.44  $\pm$ 0.82 & 65.32  $\pm$ 0.70 & 54.48  $\pm$ 0.93 & 71.32  $\pm$ 0.78\\
\hdashline
\textbf{Our STANet}  &Conv4 & \textbf{57.32 $\pm$ 0.47} &\textbf{73.00 $\pm$ 0.37} &\textbf{59.65 $\pm$ 0.51} &\textbf{74.45 $\pm$ 0.41} \\
\hline
CAN \cite{hou2019cross} &ResNet-12 &63.85 $\pm$ 0.48 & 79.44 $\pm$ 0.34 &69.89 $\pm$ 0.51 &84.23 $\pm$ 0.37 \\
P-Transfer \cite{zhiqiang2021partial} &ResNet-12 &64.21 $\pm$ 0.77 & 80.38 $\pm$ 0.59 &- &- \\
MetaOpt+ArL \cite{hongguang2021rethink} &ResNet-12 &65.21 $\pm$ 0.58 &80.41 $\pm$ 0.49 &- &-\\
DeepEMD \cite{zhang2020deepemd} &ResNet-12 &65.91 $\pm$ 0.82 & 82.41 $\pm$ 0.56 &71.16 $\pm$ 0.87 &86.03 $\pm$ 0.58 \\
IENet \cite{rizve2021exploring} &ResNet-12 &66.82 $\pm$ 0.80 & 84.35 $\pm$ 0.51 &71.87 $\pm$ 0.89 &86.82 $\pm$ 0.58 \\
DANet \cite{xu2021learning} &ResNet-12 &67.76 $\pm$ 0.46 & 82.71 $\pm$ 0.31 &71.89 $\pm$ 0.52 &85.96 $\pm$ 0.35 \\
MCL \cite{liu2022learning} &ResNet-12 &67.36 $\pm$ 0.20 & 83.63 $\pm$ 0.20  &71.76 $\pm$ 0.20 &86.01 $\pm$ 0.20 \\
Meta-DeepBDC \cite{jiangtao2022joint} &ResNet-12 &67.34 $\pm$ 0.43 & 84.46 $\pm$ 0.28 &72.34 $\pm$ 0.49 &\textbf{87.31 $\pm$ 0.32} \\
\hdashline
\textbf{Our STANet}  &ResNet-12 & \textbf{69.84 $\pm$ 0.47} &\textbf{84.88 $\pm$ 0.30} &\textbf{73.08 $\pm$ 0.49} &86.80 $\pm$ 0.34 \\
\hline
PSST \cite{zhengyu2021pareto} &WRN-28 &64.16 $\pm$ 0.44 & 80.64 $\pm$ 0.32 &- &- \\
FEAT \cite{ye2020few} &WRN-28 &65.10 $\pm$ 0.20 & 81.11 $\pm$ 0.14 &70.41 $\pm$ 0.23 &84.38 $\pm$ 0.16 \\
DANet \cite{xu2021learning} &WRN-28 &67.84 $\pm$ 0.46 & 82.74 $\pm$ 0.31 &72.18 $\pm$ 0.52 &86.26 $\pm$ 0.35 \\
\hdashline
\textbf{Our STANet}  &WRN-28 & \textbf{69.86 $\pm$ 0.46} & \textbf{85.16 $\pm$ 0.29} & \textbf{74.41 $\pm$ 0.50} &\textbf{87.64 $\pm$ 0.33} \\
\hline
\end{tabular}
\caption{Comparing to related methods on 5-way few-shot classification on miniImageNet and tieredImageNet.}
\label{table:SOTA}
\end{table*}

\subsection{Multi-Task Loss}
The three $\{f_M,f_G,f_R\}$ classifiers category the query image into the corresponding $\{N,C^{base},B\}$ classes, where $B$ is the four kinds of rotation angle. Their loses are denoted as $\{\mathcal{L}_M,\mathcal{L}_G,\mathcal{L}_R\}$, which are computed by the cross-entropy function between the corresponding predictions $\{Y_M,Y_G,Y_R\}$ and ground truth categories $\{N,C^{base},B\}$.
Then, inspired by \cite{lai2022rethinking}, the overall multi-task loss is:
\begin{small}
\begin{equation}
\begin{aligned}
\mathcal{L} = \frac{1}{2}{\mathcal{L}_M} + \sum_{j=G,R}\left({\left({\lambda}+{w_j}\right)}{\mathcal{L}_j}+{log{\frac{1}{{({\lambda}+{w_j})}}}}\right),
\end{aligned}
\label{equ:Loss}
\end{equation}
\end{small}where, ${w}$=$\frac{1}{2{\alpha^2}}$, ${\alpha}$ is a learnable weight, ${\lambda}$ is a hyper-parameter for the balance of few-shot and auxiliary tasks.
The ablation study results of ${\lambda}$ is presented in Tab.~\ref{table:ablation_mtl}.

\section{Experiment}
\subsection{Datasets and Setting}
\label{sec:Datasets}
\noindent\textbf{Datasets and Evaluation} \
Following \cite{hou2019cross,xu2021learning}, two popular benchmark datasets {\emph{mini}ImageNet} and {\emph{tiered}ImageNet} are selected, both of which are sampled from ImageNet \cite{krizhevsky2012imagenet}, and the image size is 84${\times}$84 pixels.
{\emph{mini}ImageNet} dataset contains 100 categories with 600 images per class, which is separated into $\{64,16,20\}$ categories for \{train,validation,test\} respectively.
{\emph{tiered}ImageNet} dataset has 608 categories with an average of 1281 images per class, which is divided into $\{351,97,160\}$ categories for \{train,validation,test\} respectively.
We evaluate our approaches on $5$-way $1$-shot and $5$-shot classification settings, and report \textit{average accuracy} and $95\%$ \textit{confidence interval} on $2000$ episodes randomly sampled from the novel test dataset.

\noindent\textbf{Implementation} \
The hyper-parameter $\lambda$ in Eq.~\ref{equ:Loss} is set to $1.0$ according to the results in Tab.~\ref{table:ablation_mtl}.
More implementation details of STANet for {\emph{mini}ImageNet} and {\emph{tiered}ImageNet} are referred to our released code, such as data augmentation, training epochs, optimizer and learning rate.

\subsection{Comparing to State-of-the-art Methods}
Tab.~\ref{table:SOTA} shows the comparison results between our STANet and the related FSL methods on miniImageNet and tieredImageNet, and more comparisons on CIFAR-FS is presented in APPENDIX.
Our STANet outperforms the existing SOTAs, which demonstrates the strength of our method.
Different from existing metric-based methods \cite{zhang2020deepemd,liu2022learning,jiangtao2022joint} extracting support and query features independently, our STANet can enhance the target object regions and obtains more discriminative representations.
Comparing to the state-of-the-art metric-based Meta-DeepBDC \cite{jiangtao2022joint}, STANet achieves $2.50\%$ higher accuracy on 1-shot task on miniImageNet showing the effectiveness of our STA and NTA modules.
Some metric-based methods \cite{xu2021learning,hou2019cross} integrating cross-attention module, and the proposed STANet still outperforms the competitive DANet \cite{xu2021learning} with a large improvement up to $2.08\%$, which demonstrates the strength of our Semantic and Target Attentions.

\subsection{Model Analysis}
\noindent\textbf{Influence of STA} \ \
The results in Tab.~\ref{table:ablation_1} shows that:
(I) Comparing to ProtoG, our STANet (STA) achieves consistent improvements on 1/5-shot classification tasks, because our method is able to enhance the target object regions and produce more discriminative representations via using the proposed Semantic and Target Attentions.
(II) Comparing to CAN \cite{hou2019cross} adopting the CNN-based cross attention, our SpatialFormer based STANet (SFSA, SFTA and STA) achieve obvious improvements.
(III) Comparing to ProtoG and CAN, LoFTR and SuperGlue approaches applying Transformer based cross attention, show performance degradation, which indicates that Transformer fuses different instance-level objects may cause serious disturbance on feature updating in few-shot classification task. Benefiting from the designed SpatialFormer integrating features of same instance-level objects, STANet (SFSA) obtains large accuracy improvements than LoFTR and SuperGlue.
(V) The proposed SFSA and SFTA are able to find the mutual similar object and obtain target-aware information, respectively. With their combination, the derived STA module achieves further improvements via enhancing the target object while alleviating the background distraction.

\renewcommand{\tabcolsep}{6.5pt}
\begin{table}[t]
\centering
\begin{tabular}{l | c | c  c}
\hline
\multirow{2}*{Model}  & \multirow{1}*{Attention} & \multicolumn{2}{c}{miniImageNet} \\
\cline{3-4}
 &Module & 1-shot &5-shot \\
\hline
ProtoG &None &61.80 $\pm$ 0.46 &78.59 $\pm$ 0.34  \\
CAN &CAM &63.85 $\pm$ 0.48 &79.44 $\pm$ 0.34  \\
\hline
\multirow{5}*{STANet} &LoFTR &58.30 $\pm$ 0.47 &72.06 $\pm$ 0.36  \\
 &SuperGlue &61.99 $\pm$ 0.47 &73.08 $\pm$ 0.36  \\
 &SFSA &{67.78 $\pm$ 0.47} &{82.05 $\pm$ 0.32}  \\
 &SFTA &{67.54 $\pm$ 0.47} &{82.45 $\pm$ 0.32}  \\
 &STA &\textbf{68.80 $\pm$ 0.46} &\textbf{83.12 $\pm$ 0.31}  \\
\hline
\end{tabular}
\caption{The 5-way classification results studying the influence of STA module with ResNet-12 backbone. In line with the setting of CAN \cite{hou2019cross}, STANet only applies Metric Classifier and Global Classifier (i.e. the Rotation Classifier and Novel Classifier are not adopted). Based on ProtoNet \cite{snell2017prototypical}, ProtoG adds Global Classifier to co-train the model.}
\label{table:ablation_1}
\end{table}

\renewcommand{\tabcolsep}{2.7pt}
\begin{table}[t]
\centering
\begin{tabular}{ c | c | c c | c c}
\hline
\multicolumn{1}{c|}{\multirow{2}*{$(f,r)$}}  & \multirow{2}*{Module} & \multicolumn{2}{c|}{miniImageNet}  &\multicolumn{2}{c}{tieredImageNet} \\
\cline{3-6}
\multicolumn{1}{c|}{ } & & 1-shot &5-shot &1-shot &5-shot \\
\hline
$(Q,P)$  &Transformer &60.60  &74.13 &66.17 &79.79 \\
$(Q,P)$  &SpatialFormer &\textbf{67.78} &\textbf{82.05} &\textbf{71.45} &\textbf{84.73} \\
\hline
$(Q,W_G)$  &Transformer &{67.39}  &{82.02} &{71.09} &{84.54} \\
$(Q,W_G)$  &SpatialFormer &\textbf{67.54} &\textbf{82.45} &\textbf{71.71} &\textbf{85.24} \\
\hline
\end{tabular}
\caption{The 5-way classification results studying the influence of SpatialFormer with ResNet-12 backbone. The setting is the same as Tab.~\ref{table:ablation_1}.}
\label{table:ablation_former}
\end{table}

\noindent\textbf{Influence of SpatialFormer} \ \
As shown in Tab.~\ref{table:ablation_former}, in the setting of assigning $(f,r)$ with $(Q,P)$ and $(Q,W_G)$, the proposed SpatialFormer achieves consistent improvements on 1/5-shot classification tasks on mini/tieredImageNet, which shows the superiority of our SpatialFormer than Transformer.
Besides, comparing to the setting of $(f,r)=(Q,P)$, applying $(f,r)=(Q,W_G)$ obtains higher performance, which indicates that the class weights $W_G$ learned from the base dataset are useful in few-shot learning.

\noindent\textbf{Influence of Multi-Task Loss} \ \
As shown in Tab.~\ref{table:ablation_mtl}, our STANet achieves the optimal results with ${\lambda}$=${1.0}$, which obtains a large accuracy improvements on 1/5-shot tasks than without Rotation Classifier. These results also indicate that the rotation task is effective for performance improvement due to learning a more representative embedding model.

\renewcommand{\tabcolsep}{2.3pt}
\begin{table}[t]
\centering
\begin{tabular}{c | c c c | c  c | c  c}
\hline
\multirow{2}*{${\lambda}$} & \multicolumn{3}{c|}{Loss weights} & \multicolumn{2}{c|}{ResNet-12} & \multicolumn{2}{c}{WRN-28} \\
\cline{2-8}
&Metric &Global &Rotation & 1-shot &5-shot & 1-shot &5-shot \\
\hline
-&0.5 &1.0 &- &68.80 &83.12 &67.23 &82.27 \\
-&0.5 &1.0 &1.0 &{68.97} &{83.70} &{68.93} &{84.14} \\
\hline
0.5&0.5 &${w_G}$ &${w_R}$ &{69.42} &{83.74} &69.33 &83.87 \\
1.0&0.5 &${w_G}$ &${w_R}$ &\textbf{69.48} &\textbf{84.12} &\textbf{69.55} & \textbf{84.45} \\
1.5&0.5 &${w_G}$ &${w_R}$ &69.29 &83.29 &{69.55} &{83.94}  \\
\hline
\end{tabular}
\caption{The 5-way classification results of STANet on \emph{mini}ImageNet studying the influence of multi-task loss.}
\label{table:ablation_mtl}
\end{table}

\noindent\textbf{Influence of NTA} \ \
As illustrated in Tab.~\ref{table:ablation_NCA2}, the results under different backbone and dataset show that our NTA module consistently obtains accuracy improvements on 1/5/10-shot tasks, which indicates the proposed NTA module is effective.
With the increase of shots, the NTA module obtains larger accuracy improvement, which is due to the increase of support data which leads to learned a more representative weights $W$ of Novel Classifier.

\renewcommand{\tabcolsep}{2.0pt}
\begin{table}[t]
\centering
\resizebox{8.5cm}{!}
{
\begin{tabular}{ c | c | c c c | c c c}
\hline
\multicolumn{1}{c|}{\multirow{2}*{NTA}}  & \multirow{2}*{B.b.} & \multicolumn{3}{c|}{miniImageNet}  &\multicolumn{3}{c}{tieredImageNet} \\
\cline{3-8}
\multicolumn{1}{c|}{ } & & 1-shot & 5-shot &10-shot  & 1-shot &5-shot &10-shot \\
\hline
{-}  &R.12 &{69.48} &{84.12}  &86.91 &72.70 &86.28 &88.89 \\
{\checkmark}  &R.12 &\textbf{69.84} &\textbf{84.88}  &\textbf{87.98} &\textbf{73.08} &\textbf{86.80} &\textbf{89.60} \\
\hline
{-}  &W.28 &{69.55} & {84.45} &86.97  &74.01 &87.04 &89.41 \\
{\checkmark}  &W.28 &\textbf{69.86} & \textbf{85.16} &\textbf{88.09} &\textbf{74.41} &\textbf{87.64} &\textbf{90.10} \\
\hline
\end{tabular}
}
\caption{The 5-way classification results of STANet studying the influence of NTA, with ${\lambda}$=$1.0$.
The B.b., R.12 and W.28 represent Backbone, ResNet-12 and WRN-28, respectively.}
\label{table:ablation_NCA2}
\end{table}

\begin{figure}[!t]
\centering
\includegraphics[width=0.9\linewidth]{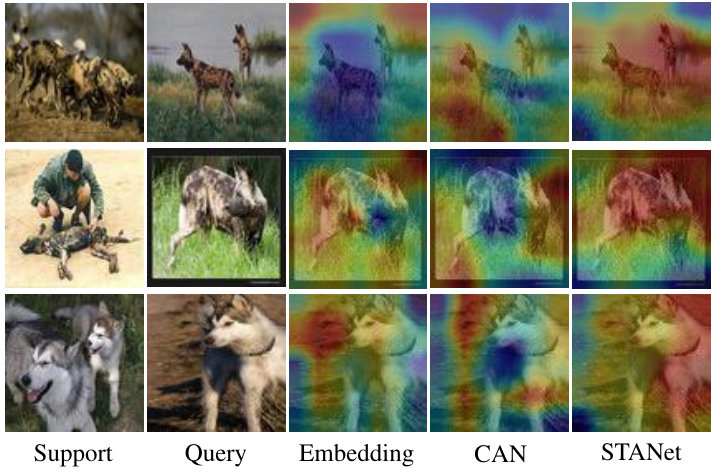}
\caption{The class activation maps generated from 5-way 5-shot task, where \textit{Embedding} means the features extracted after the backbone in STANet.}
\label{fig:visual}
\end{figure}

\subsection{Visualizations}
Fig.~\ref{fig:visual} shows the class activation maps \cite{zhou2016learning} of CAN and our STANet. By comparing STANet to \textit{Embedding}, it shows that STANet is able to highlight the target object which is unseen in the pre-training stage. Comparing to CAN, our STANet is more accurate and has larger receptive field. Obviously, CAN suffers the background distraction problem according to its class activation maps which focus on the background. With the proposed STA and NTA modules, our STANet obtains global attention preventing the target losing in the background, and decreases the background noise to help the model focus on the target objects.
More visualizations are presented in APPENDIX.

\section{Conclusion}
In this paper, we proposed a Semantic and Target Attentions Network (STANet) for few-shot classification.
Firstly, a novel SpatialFormer structure is designed for few-shot learning, which explores semantic-level similar between inputs.
Secondly, two effective modules named STA and NTA are introduced to produce discriminative representations via enhancing target object while filtering out background noise.
The NTA module shifts input feature with class weights learned from novel task to increase inter-class feature distance and involve task-aware information. Then, the SpatialFormer based STA module highlights the target object regions to generate more discriminative representations.
Extensive experiments show that the proposed approaches are effective, and our STANet achieves new state-of-the-arts on multiple few-shot classification benchmark datasets.

\section{Acknowledgments}
This work was supported by the National Key Research and Development Program of China (2021ZD0111000), National Natural Science Foundation of China No. 62222602 and No. 62176092, Shanghai Science and Technology Commission No.21511100700, Natural Science Foundation of Shanghai (20ZR1417700).

\bibliography{aaai23}

\newpage
\section{Algorithm of STANet}
The algorithm of STANet is presented in Algorithm~\ref{alg:episode}.
The STANet contains six parts: embedding backbone ${f_\theta}$, Semantic and Target Attentions (STA) ${f_{STA}}$ module, Metric ${f_M}$ and Novel ${f_N}$ few-shot classifiers, and auxiliary Rotation ${f_R}$ and Global ${f_G}$ classifiers.
The training and inference of STANet can be divided into two steps: \textbf{Step 1}, train the base model ${f_{base}={[f_\theta,f_{STA},f_M,f_G,f_R]}}$ on the base set ${X^{base}}$; \textbf{Step 2}, fine-tune the Novel Classifier ${f_N}$ and make predictions with ${f_{MN}={[f_\theta,f_{STA},f_M,f_N]}}$ on the novel set ${X^{novel}}$.

\begin{algorithm}
\caption{STANet training and inference}
\label{alg:episode}
\SetAlgoLined
\SetKwInput{KwData}{Input}
\SetKwInput{KwModel}{Model}
\SetKwInput{KwResult}{Output}
 \KwData{$(\mathcal{S},\mathcal{Q}) \in {X^{base}}$; $(\mathcal{S{'}},\mathcal{Q{'}}) \in {X^{novel}}$; training and testing epochs $E_{1}$ and $E_{2}$}
 \KwModel{${f_\theta}$; ${f_{STA}}$; ${f_M}$; ${f_G}$; ${f_R}$; ${f_N}$}
 \KwResult{Inference accuracy $Acc$ on ${X^{novel}}$}
 \Begin{
 \textbf{Step 1:} Model training on base set.

 \For{i \textbf{from} 1 \textbf{to} $E_{1}$}{
 Sample training data $(\mathcal{S},\mathcal{Q}) \in {X^{base}}$\;
 Calculate loss $\mathcal{L}=\mathcal{L}_M+\mathcal{L}_G+\mathcal{L}_R$\;
 Optimize [${f_\theta}$,${f_{STA}}$,${f_M}$,${f_G}$,${f_R}$] with SGD\;
  }

 \textbf{Step 2:} Fine-tuning and inference on novel set.
 Freeze [${f_\theta}$,${f_{STA}}$,${f_M}$,${f_G}$,${f_R}$]; $acc$ = [];

 \For{j \textbf{from} 1 \textbf{to} $E_{2}$}{
 Sample testing data $(\mathcal{S{'}},\mathcal{Q{'}}) \in {X^{novel}}$\;
 Calculate cross entropy loss $\mathcal{L}_{N}$ for ${\mathcal{S{'}}}$\;
 Optimize ${f_N}$ with SGD\;
 Update the output of ${f_\theta}$ by NTA\;
 Predict $\mathcal{Q{'}}$\ with model [${f_\theta}$,${f_{STA}}$,${f_M}$,${f_N}$]\;
 Calculate accuracy and append it to $acc$\;
  }
calculate accuracy $Acc$\ with $acc$\;
\textbf{return}  $Acc$.}
\end{algorithm}

\renewcommand\thetable{6}
\renewcommand{\tabcolsep}{2.0pt}
\begin{table}[ht]
\caption{Comparison on 5-way classification task on CIFAR-FS dataset with ResNet-12 backbone.}
\centering
\begin{tabular}{ l | c | c c}
\hline
\multicolumn{1}{l|}{\multirow{2}*{Model}}  & \multicolumn{2}{c}{CIFAR-FS} \\
\cline{2-3}
\multicolumn{1}{c|}{ } & 1-shot &5-shot \\
\hline
RFS~\cite{tian2020rethinking} &71.50 $\pm$ 0.80 &86.00 $\pm$ 0.50 \\
MetaOpt~\cite{lee2019meta} &72.60 $\pm$ 0.70 &84.30 $\pm$ 0.50 \\
DSN-MR~\cite{simon2020adaptive} &75.60 $\pm$ 0.90 &86.20 $\pm$ 0.60 \\
IENet~\cite{rizve2021exploring} &76.83 $\pm$ 0.82 &89.26 $\pm$ 0.58 \\
\hdashline
\textbf{Our STANet} &\textbf{79.53 $\pm$ 0.47} &\textbf{89.87 $\pm$ 0.32} \\
\hline
\end{tabular}
\label{table:SOTA_cifar}
\end{table}

\renewcommand\thefigure{6}
\begin{figure}[ht]
\centering
\includegraphics[width=0.9\linewidth]{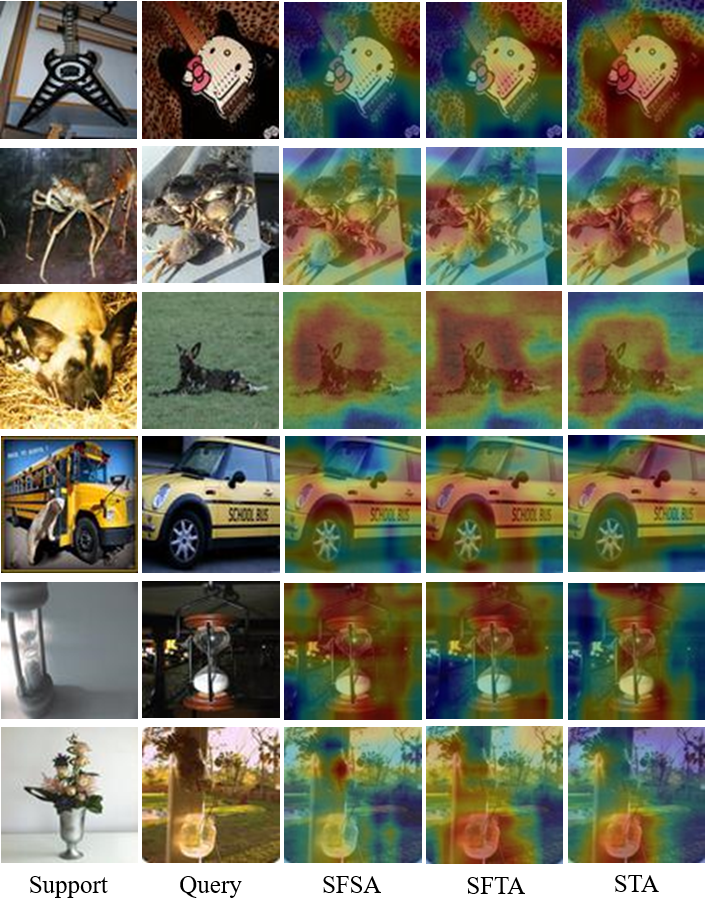}
\caption{The visualizations of SFSA, SFTA and STA.}
\label{fig:visual_compair}
\end{figure}

\renewcommand\thefigure{7}
\begin{figure}[ht]
\centering
\includegraphics[width=0.99\linewidth]{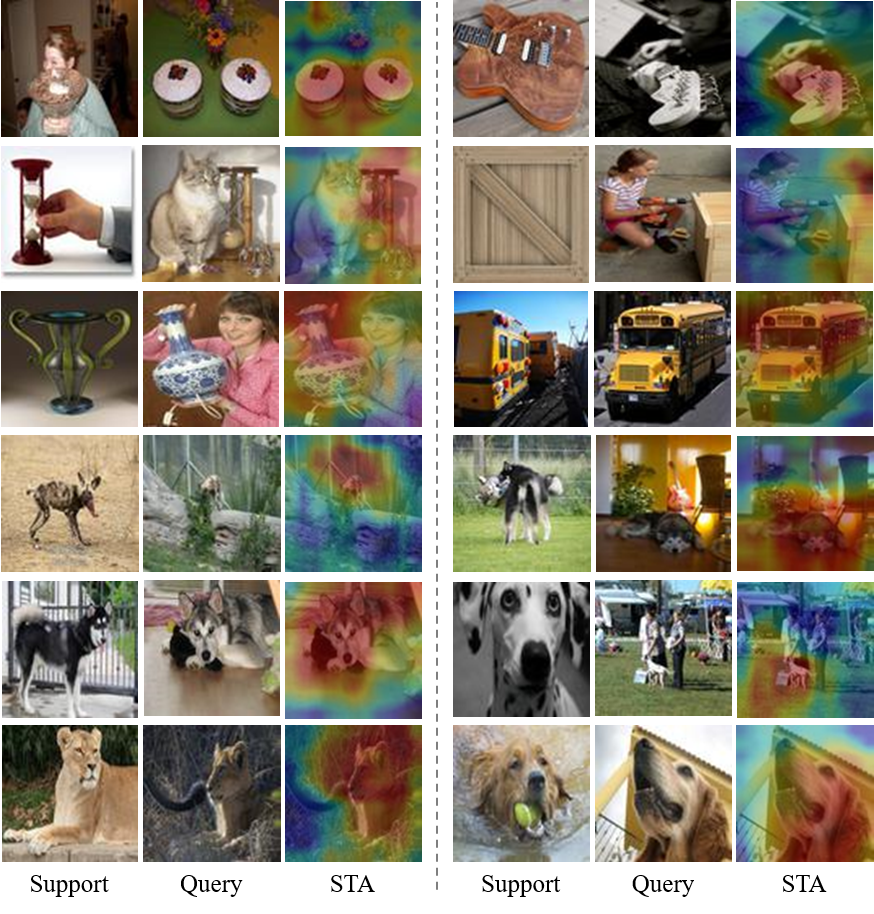}
\caption{The visualizations of STA.}
\label{fig:visual_STA}
\end{figure}

\section{Comparison with SOTAs on CIFAR-FS}
CIFAR-FS dataset is constructed by randomly splitting the 100 classes of the CIFAR-100 dataset into 64, 16, and 20 categories for train, validation, and test.
Tab.~\ref{table:SOTA_cifar} shows that our STANet outperforms the existing SOTAs on CIFAR-FS dataset, which demonstrates the strength of our approach.

\section{Visualizations}
The visualizations of SFSA, SFTA and STA are shown in Fig.~\ref{fig:visual_compair} and Fig.~\ref{fig:visual_STA}.
The proposed SFSA and SFTA modules, are utilized to highlight the mutual similar object and obtain target-aware information, respectively. With their combination, the derived STA module is able to enhance target object while alleviate background distraction.

\section{More Discussions}
\subsection{More Explanations for \emph{Problem \ding{172}} and \emph{Problem \ding{173}}}
(I) \emph{Problem \ding{172}} means that the current CNN-based method models correlation between pairs based on local features, which produces inaccurate attention maps due to the target object is located randomly with different scale among the image.
(II) \emph{Problem \ding{172}} focuses on how to better measure the simlarity between pair features, while \emph{Problem \ding{173}} points out the distraction of similar backgrounds, i.e. even if \emph{Problem \ding{172}} is perfectly solved, \emph{Problem \ding{173}} is still exist.

\subsection{Parameters Comparison}
(I) The parameters of SpatialFormer and Transformer are the same since SpatialAttention is a non-parameter operator. As illustrated in Tab.~2, Our SpatialFormer-based SFSA and SFTA obtain large accuracy improvements than Transformer-based LoFTR and SuperGlue.
(II) The parameters with backbones of \{ResNet12, WRN28\} are listed as follows: ProtoG = \{7.75M, 35M\}, LoFTR = SuperGlue = SFSA = SFTA = \{9.25M, 36.5M\}, STA=\{10.75M, 38M\}.
The parameters of a single SpatialFormer layer is 1.5M. Removing $\{W_Q,W_K,W_V\}$ reduces the parameters of SpatialFormer into 1M, while causes very slight influence on accuracy.

\subsection{Further Applications of SpatialFormer}
(I) We have further explored the applications of our SpatialFormer on few-shot segmentation and object detection tasks, and have achieved competitive improvements.
(II) In detail, we further derive a SpatialFormer Embedding Attention (SFEA), via replacing the classifier weights $W_G$ of SFTA in Eq.~(7) to the learnable query embedding $W_E$ (pytorch code is $W_E$ = $nn.Embedding(C^{base}, c)$). Then, the SFEA can enhance the target regions of input features that are semantically similar to $W_E$, i.e. $f' = f_{SFEA}(f) = SpatialFormer(f,W_E)$.
(III) Stacking the SFEA after the backbone, we achieve improvements on: (a) PASCAL-5i segmentation: RePRI+ResNet50 (59.3/64.8 on 1/5 shot) vs. +SFEA (61.1/65.9). (b) VOC (Novel Set 1) detection with G-FSOD setting: DeFRCN+ResNet101 (40.2/63.6 on 1/5 shot) vs. +SFEA (43.1/65.7).

\end{document}